\documentclass[sigconf,nonacm,balance=false]{acmart-liveasset}
\usepackage{amsmath,tabularx,tikz}
\usetikzlibrary{arrows.meta,positioning,fit}
\setcopyright{none}
\title{LiveAsset: Learning Intent Assets from Corrective Interactions for Agent Runtime Control}
\author{Guanming Liu}
\affiliation{%
  \institution{Fudan University}
  \country{}
}
\author{Weihang Wang}
\affiliation{%
  \institution{Fudan University}
  \country{}
}
\author{Peng Zhang}
\affiliation{%
  \institution{Fudan University}
  \country{}
}
\author{Meng Wu}
\affiliation{%
  \institution{Lenovo Group Ltd}
  \country{}
}
\author{Kainan Tu}
\affiliation{%
  \institution{Fudan University}
  \country{}
}
\author{Jiahao Liu}
\affiliation{%
  \institution{Fudan University}
  \country{}
}
\author{Yu Zhang}
\affiliation{%
  \institution{Lenovo Group Ltd}
  \country{}
}
\author{Liuxin Zhang}
\affiliation{%
  \institution{Lenovo Group Ltd}
  \country{}
}
\author{Qianying Wang}
\affiliation{%
  \institution{Lenovo Group Ltd}
  \country{}
}
\author{Ning Gu}
\affiliation{%
  \institution{Fudan University}
  \country{}
}
\author{Tun Lu}
\affiliation{%
  \institution{Fudan University}
  \country{}
}

\begin{document}
\begin{abstract}
Large language model (LLM) agents acting across web applications and online services often need multi-turn corrections to align their work with users' underspecified goals and preferences. In practice, users may identify problems in the current outcome without specifying the desired result, so learning user intentions requires interpreting each correction against the agent's preceding and revised outcomes. These expectations are often connected to a particular context, making it hard to know when they apply and how to ensure the agent later acts as the user expects.
To this end, we propose LiveAsset, an intent harness that learns reusable intent assets from corrective interactions and uses them for runtime control of agent execution. LiveAsset jointly interprets each correction with the agent's original and revised results to infer user intent, incrementally updating the assets while preserving earlier requirements that remain applicable. The learned requirements guide agent execution from task entry and provide criteria for runtime verification, which uses the explanations to assess applicability before each reply is delivered and directs the agent to revise the result to address unmet requirements while preserving those already satisfied.
Following a learning phase in which simulated users progressively reveal requirements through multi-turn corrections, we evaluate LiveAsset using same-task replay of the original requests on $\pi$-Bench, PERMA, and Vita and cross-task transfer on PAST-Bench and SkillEvolBench. Experiments show that LiveAsset better fulfills requirements learned from corrective interactions and improves performance on these benchmarks compared with memory, skill, and rule-based harness baselines. Code is available at \url{https://github.com/landian60/LiveAsset}.
\end{abstract}
\maketitle
\begin{figure}[!ht]
\centering
\includegraphics[width=\columnwidth]{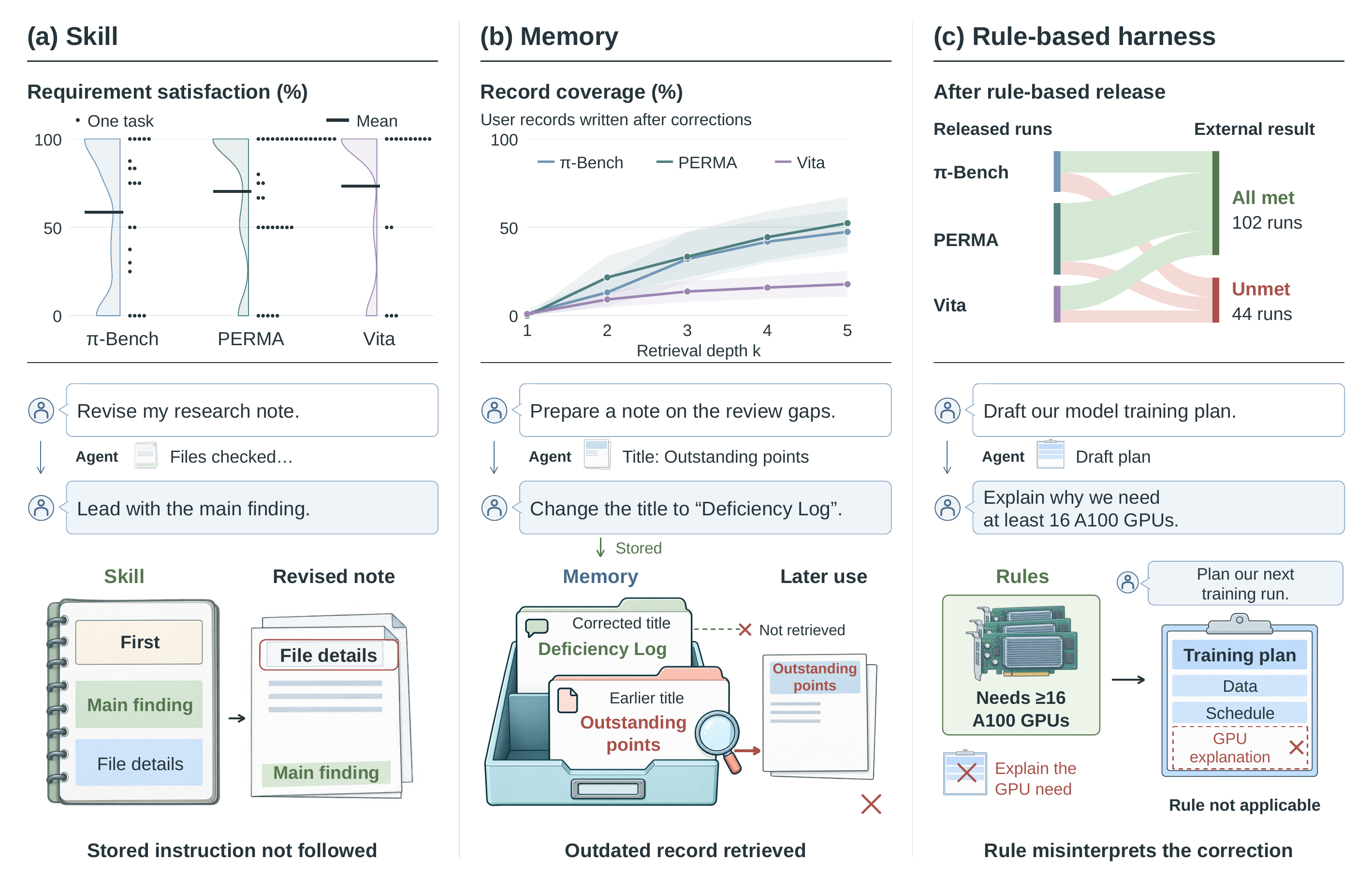}
\caption{Challenges for existing approaches in learning from corrective interactions in human--agent collaboration. The top panels report (a) requirement satisfaction with stored skills, (b) retrieval coverage of memory records written after corrections, and (c) requirement fulfillment after rule-based release, and the bottom panels illustrate the corresponding failures.}
\label{fig:motivation}
\end{figure}

\section{Introduction}
\label{sec:intro}
LLM agents allow users to delegate tasks across applications and online services through natural-language requests. The initial request rarely specifies every condition the outcome must satisfy, since users' expectations depend on personal preferences and the current situation~\cite{yang2026underspecification}, and users often settle their criteria only after seeing outputs~\cite{shankar2024validators}. As users review the agent's work over several turns, they clarify or revise their preferences and task requirements, and the agent needs to incorporate this feedback into its next decisions and actions. Each correction has to be understood alongside earlier feedback, because changing one aspect of the work does not necessarily replace the user's other expectations. Learning from these exchanges could help the agent infer user intent and use it to guide decisions and actions in similar tasks.

Existing approaches store interaction experience as reusable skills~\cite{lei2026skillevolbench}, memories~\cite{chhikara2025mem0}, or constraints checked during execution~\cite{zhou2026trace,wang2026agentspec,qin2026statem}. Learning from corrective interactions involves understanding how the agent's work differs from the user's expectations and what the agent should change, even when the feedback leaves the desired result implicit. In our analysis of coding dialogues from SWE-chat~\cite{baumann2026swechat}, 135 of 331 corrective or clarifying messages identify a discrepancy without stating a requirement for the desired result (Appendix~\ref{app:feedback-analysis}). The preceding outcome helps locate the discrepancy, while the subsequent revision reveals which choices changed and how those changes may address the user's feedback, providing evidence for inferring the expected result. These expectations are often connected to a particular context, so later execution requires judging when the learned intent applies and ensuring that the agent acts as the user expects. Even when prior preferences or knowledge are available, agents can overlook relevant guidance or fail to act on it~\cite{zhao2025prefeval,schmied2026greedy}. Figure~\ref{fig:motivation} pairs benchmark diagnostics with illustrative task examples: a stored instruction is not followed, an outdated record is retrieved despite a stored correction, and a compiled rule misinterprets the user's correction. We investigate how agents can learn user intent from corrective interactions and use it to guide later execution and verify its outcomes.

We propose LiveAsset, an intent harness that uses the learned user intents to guide agent execution and verify whether applicable requirements are satisfied. LiveAsset learns these intents by jointly interpreting each correction with the agent's original and revised results. Concretely, LiveAsset preserves the inferred user intent in reusable assets for later tasks. The intent assets combine intent explanations that relate the user's expectations to the task context with checkable requirements for the agent's actions and outcomes. Successive corrections update the intent assets while preserving earlier requirements that remain applicable. For a new request, LiveAsset selects relevant intents to guide agent decisions and actions. Before delivery, LiveAsset checks the current result against the selected intents, using the intent explanations to assess applicability and the requirements to assess fulfillment. When applicable requirements remain unmet, LiveAsset directs the agent to revise the result and checks the revised result again.

We evaluate LiveAsset using intent assets learned from simulated users' multi-turn corrections under a reproducible protocol informed by the forms of feedback observed in SWE-chat. Same-task replay tests whether the agent fulfills learned requirements when re-executing the original request from its initial environment, while cross-task transfer tests whether it can use the learned experience on tasks with different requests or contexts. We conduct replay on $\pi$-Bench~\cite{zhang2026pibench}, PERMA~\cite{liu2026perma}, and Vita~\cite{chen2026vitabench2}, and transfer on PAST-Bench~\cite{xue2026pastbench} and SkillEvolBench~\cite{lei2026skillevolbench}. The results show improvements over memory, skill, and rule-based harness baselines in learned requirement fulfillment and performance on changed tasks. Further analyses examine how learned intents are selected at task entry, how runtime checks relate to corrective actions, and how consistently requirements are fulfilled across repeated executions.

Our main contributions are as follows.
\begin{itemize}
\item We propose LiveAsset, which learns user intent by jointly interpreting corrections and the original and revised results, preserving contextual explanations together with checkable requirements to guide and verify the agent's actions in later tasks.
\item We formulate a corrective-interaction learning setting for complex tasks, using a user simulator that progressively reveals requirements and provides feedback grounded in the agent's outcomes.
\item Experiments on five benchmarks demonstrate that LiveAsset improves requirement satisfaction and cross-task transfer performance compared with memory, skill, and rule-based harness baselines.
\end{itemize}

\section{Related Work}
\label{sec:related}
\subsection{Learning User Intent from Interaction}
Multi-turn interaction allows users to clarify their intent and refine an agent's work through feedback and corrections. Prism organizes clarification around dependencies among intent elements, and RECAP rewrites ambiguous dialogues into concise user goals for agent planning~\cite{liao2026prism,mitra2026recap}. Naturally occurring conversational feedback supports alignment with human preferences, and PRELUDE and PROSE learn preferences from edits and writing samples, respectively~\cite{donyehiya2025feedback,gao2024prelude,aroca2025prose}, while PAHF and MultiSessionCollab update preferences from feedback within and across interaction sessions~\cite{liang2026pahf,mehri2026multisessioncollab}. On the Web, PUMA, Persona2Web, and MemWeaver personalize agent behavior using historical user behaviors and preferences, and AgenticShop evaluates personalized shopping against user-specific checklists~\cite{cai2025personalwab,kim2026persona2web,yu2026memweaver,kim2026agenticshop}. However, a correction to an agent's intermediate results may identify a discrepancy while leaving the desired result implicit. LiveAsset learns reusable intents from successive corrections and agent revisions to capture users' evolving expectations.

\subsection{Reusing Experience in Agent Execution}
Agent harnesses organize the information and execution controls around an LLM agent. Skills and memory supply reusable guidance within this process, as SkillEvolBench, Mem0, ExpeL, and MetaKube reuse procedural skills, conversational information, task insights, and diagnosis patterns, respectively~\cite{lei2026skillevolbench,chhikara2025mem0,zhao2024expel,sun2026metakube}. Skills and memories can encode conditional requirements, and rule-based harnesses further check such requirements during execution. TRACE compiles user corrections into rules with applicability checks and runtime enforcement, implemented in TellOnce~\cite{zhou2026trace,tellonce2026}, while AgentSpec and StateM constrain execution through triggers and predicates or checked state transitions~\cite{wang2026agentspec,qin2026statem}. IDSS tracks user intents, tool-grounded facts, and constraints in an explicit situation state within a multi-turn task~\cite{tao2026idss}. Reused experience must also be applied selectively, since LLMs often misjudge which prior context is relevant to a new request and overuse personal information when it is unnecessary~\cite{kim2025cupid,hu2026opbench}. Determining whether learned user requirements apply to a new task therefore requires interpreting the agent's actions and outcomes during execution. LiveAsset jointly interprets the results before and after each correction to learn requirements together with explanations of their context and rationale, which remain available during later execution to judge applicability and guide verification and revision.

\section{LiveAsset}
\label{sec:method}
LiveAsset learns reusable intent assets from corrective interactions to guide the agent's actions and verify their outcomes in later tasks (Figure~\ref{fig:overview}). It interprets each correction alongside the original and revised results, expressing the user's expectations through an intent explanation of their context and rationale, checkable requirements, and related facts. We first formulate the learning and reuse problem, then describe how these assets are incrementally learned, selected for a new task, and used for runtime verification.

\subsection{Problem Formulation}
\label{sec:problem}
\subsubsection{Corrective Interaction Setting}
\label{sec:feedback-analysis}
\label{sec:simulation}
We consider a multi-turn interaction in which a user submits a task request, reviews the agent's work, and provides feedback that clarifies or revises the requirements. The user may state a desired change directly or identify a discrepancy without specifying how the result should change, leaving the agent to interpret the feedback and revise the result while continuing to follow earlier requirements that remain applicable. Successive results and corrections progressively reveal the user's expectations and form learning histories.

\begin{figure}[t]
\centering
\includegraphics[width=\linewidth]{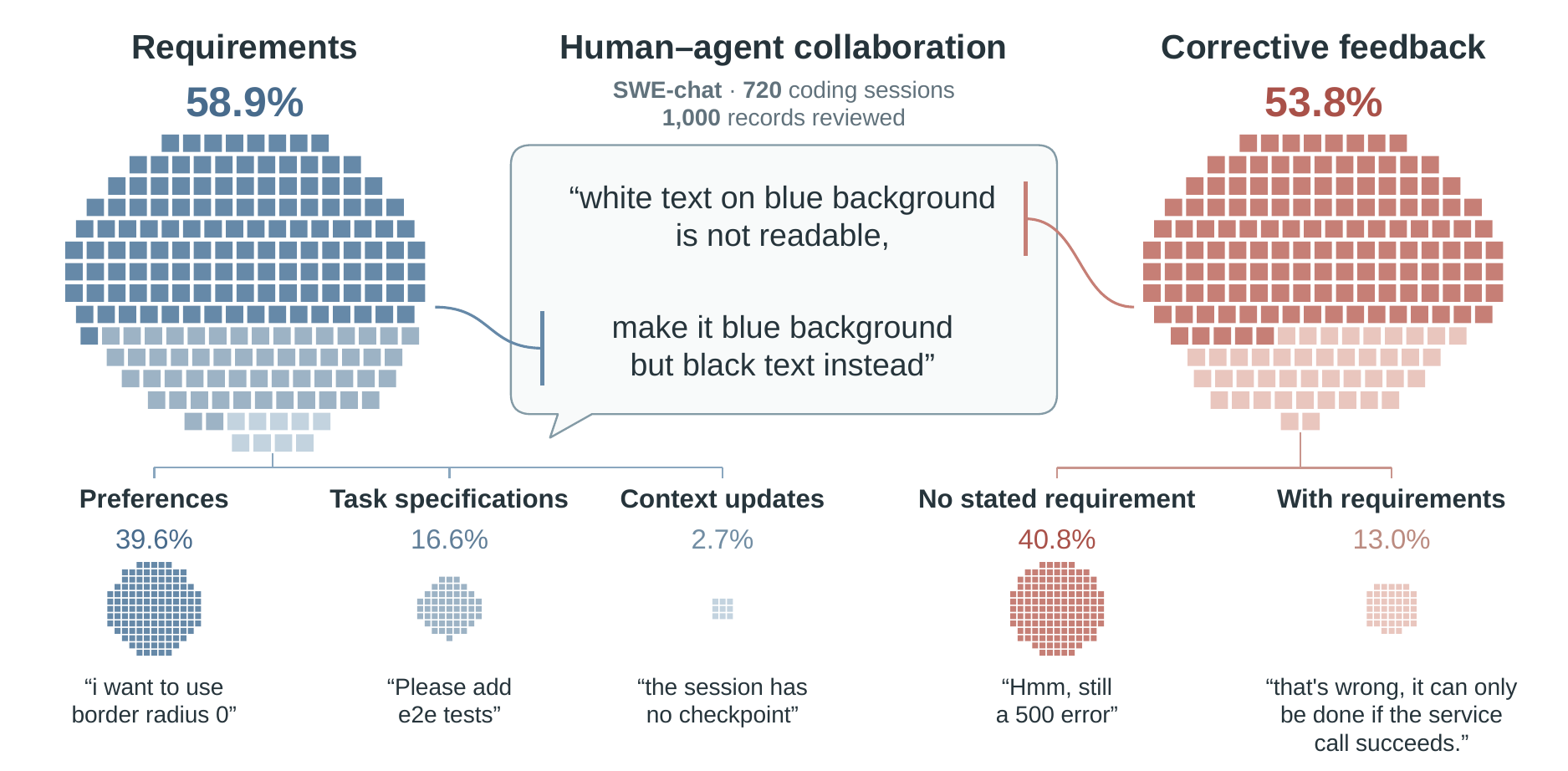}
\caption{Feedback forms in human--agent collaboration, analyzed to inform the design of simulated corrective interactions. Stated requirements (left) and corrective feedback (right) are counted over 331 SWE-chat messages, with excerpts showing their forms and overlap.}
\label{fig:correction-content}
\end{figure}

We use a simulated user to generate learning interactions informed by the feedback forms observed in SWE-chat~\cite{baumann2026swechat} (Figure~\ref{fig:correction-content}). The simulator has access to task-specific user information, including hidden preferences where provided, and the task's evaluation criteria. It reviews the interaction history and current result to assess which requirements are satisfied. Preferences are disclosed progressively, and clarification questions are answered from the supplied user information. Corrective feedback usually raises one unmet requirement or a few related requirements, pointing out a concrete discrepancy where possible and stating the desired change otherwise. After each revision, the user simulator reassesses fulfillment, tracks requirements already raised, and flags previously satisfied requirements that have become unmet. Interaction continues until the user simulator accepts the result or the dialogue budget is exhausted.

The requests and feedback from user $u$, together with the agent's delivered results and visible tool observations and files, form the learning history $H_u$, which serves as the input to LiveAsset's intent learning. The user simulator's private specification, evaluation criteria, and assessment records are not included directly in $H_u$.

\begin{figure*}[t]
\centering
\includegraphics[width=0.9\textwidth]{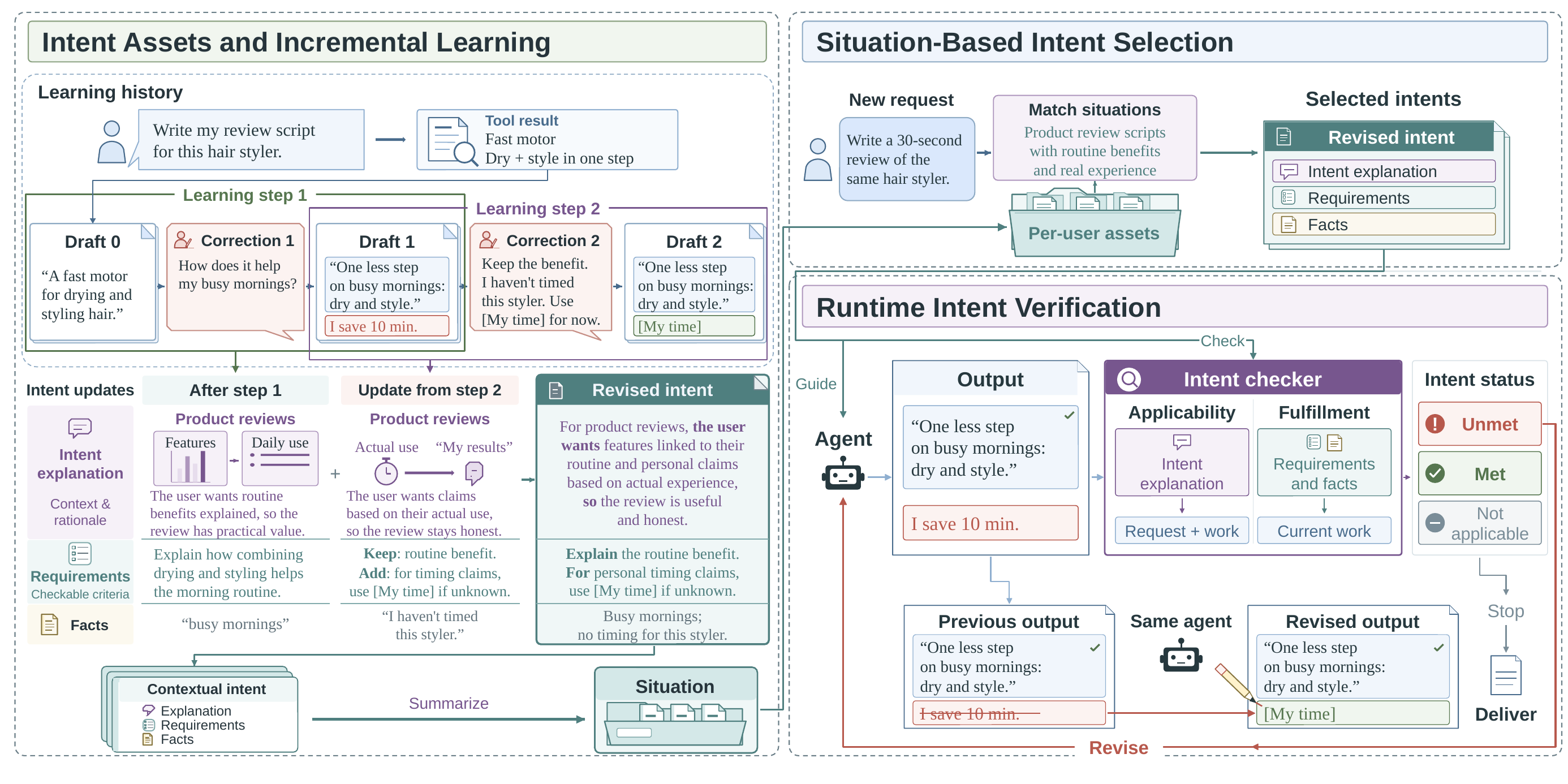}
\caption{Overview of LiveAsset. Left: incremental learning of intent assets from corrections and the original and revised results. Right: situation matching selects relevant intents, and runtime checks assess their applicability and fulfillment to guide revision by the same agent.}
\label{fig:overview}
\end{figure*}

\subsubsection{Intents and Reuse Objective}
We describe the expectations revealed in $H_u$ as a set of user intents $\mathcal{J}_u$. Each intent $j\in\mathcal{J}_u$ has an applicability condition $\alpha_j(q,y)\in\{0,1\}$ and a criterion $\beta_j(q,y)\in\{0,1\}$, defined on a request $q$ and the agent's actions and outcomes $y$ at a given point, including its reply, tool actions, and files. The condition indicates whether the intent applies, and the criterion indicates whether the intent is fulfilled. Applicability depends on both the request and the current actions and outcomes. For example, a draft that states ``I save 10 min.'' makes the requirement to base personal claims on actual experience applicable (Figure~\ref{fig:overview}). When user feedback $c_k$ identifies a discrepancy in a delivered result $x_{k-1}$, at least one applicable intent remains unfulfilled,
\begin{equation}
\exists\, j\in\mathcal{J}_u,\quad \alpha_j(q,x_{k-1})=1\ \wedge\ \beta_j(q,x_{k-1})=0.
\label{eq:correction}
\end{equation}
Although a correction indicates a mismatch between the result and the user's expectations, the criterion for resolving the mismatch may remain implicit and must be inferred from the feedback and surrounding interaction context. The preceding result grounds the correction in the actions or outcomes the user reviewed, while the subsequent revision provides evidence of how the discrepancy can be addressed, subject to refinement through further feedback. LiveAsset approximates $\alpha_j$ with the explanation of a learned intent and $\beta_j$ with its requirements and facts (Section~\ref{sec:learning}).

At test time, we evaluate whether learning from multi-turn corrective interactions improves the agent's performance on subsequent tasks. Given a target request $q$, the objective is to complete the task with actions and outcomes $y$ that satisfy the preferences and other requirements revealed in $H_u$ wherever they apply,
\begin{equation}
\alpha_j(q,y)=1\ \Rightarrow\ \beta_j(q,y)=1,\quad \forall\, j\in\mathcal{J}_u.
\label{eq:objective}
\end{equation}
We examine this objective in two settings. In \emph{same-task replay}, the agent executes the original request again from its initial environment, testing whether it can meet requirements revealed in the earlier interaction. In \emph{transfer to changed tasks}, the agent executes different, related requests, testing which requirements and practices identified in earlier interactions remain useful beyond the original task. In both settings, we evaluate the agent without the user simulation that produced $H_u$, assessing its ability to complete the target task after learning from prior multi-turn corrective interactions. Section~\ref{sec:data} gives the task-specific interaction settings and budgets.

\subsection{Intent Assets and Incremental Learning}
\label{sec:learning}
LiveAsset learns incrementally from overlapping correction steps in $H_u$, each comprising the original result, a user correction, and the subsequent revision. We denote a step by $d_k=(x_{k-1},c_k,x_k)$, where $c_k$ is the correction and $x_{k-1},x_k$ include the visible replies, tool observations, and files before and after it. Given $d_k$ and the assets produced by the preceding step, LiveAsset interprets the feedback in relation to the original and revised results, using the preceding result to identify the object of the correction and the revision to examine how the relevant actions or outcomes changed. When feedback leaves the expected result implicit, LiveAsset compares the original and revised results to infer the underlying requirements and why they matter in that context.

The inferred requirements describe the expected outcomes and the practices needed to achieve them, with the user's stated requests constraining how the agent's actions are interpreted. Tool names, file layouts, and an incidental sequence of actions are examined in terms of the purpose they serve, so that the learned requirements capture reusable expectations supported by the feedback and surrounding interaction context. As new corrections arrive, LiveAsset adds or revises intents and their explanations while preserving earlier requirements and facts that the feedback does not contradict, allowing later use to draw on the context in which the requirements arose. Figure~\ref{fig:overview} illustrates how successive corrections refine an intent while preserving earlier requirements.

A \emph{contextual intent} $i=(e_i,R_i,F_i)$ represents the user's expected outcome through an intent explanation $e_i$, a corresponding requirements list $R_i$, and a related facts list $F_i$. The \emph{intent explanation} describes the task context and rationale underlying the expected outcome, providing the semantic basis for assessing applicability to a later request and the agent's actions and outcomes during execution. The requirements specify reusable criteria that can be checked in the agent's reply, tool actions, or files, while the facts preserve concrete user-provided information with its original scope and supporting quotations.

In Figure~\ref{fig:overview}, the first correction asks how the hair styler helps on busy mornings. Comparing Draft~0 with Draft~1 connects combined drying and styling to one less step in the user's routine. The explanation $e_i$ records why that connection matters in a product review, and $R_i$ requires the script to explain the routine benefit. The second correction asks the agent to keep this explanation but rejects an unsupported personal timing claim. LiveAsset extends $e_i$ to explain why personal claims should reflect actual experience and adds a requirement to use \texttt{[My time]} when timing is unknown. The revised intent combines both expectations. The facts $F_i$ record that the user has busy mornings and has not timed this styler.

LiveAsset organizes related intents under a \emph{situation} $s$, whose associated intents are denoted by $\mathcal{I}(s)$. After processing all correction steps in a history, LiveAsset generates the situation description from the requests and intent explanations, summarizing the request context, expected outcome, and rationale for task-entry selection. For example, the illustrated intent belongs to a situation describing product review scripts that connect features to the user's routine and ground personal claims in actual experience. Situations with shared request contexts and user objectives may be merged, with their intents collected under one description.

\subsection{Situation-Based Intent Selection}
\label{sec:selection}
At task entry, LiveAsset selects situations by matching their descriptions to the request $q$ and available task context. For example, a request for a 30-second review of the same hair styler matches the situation for product review scripts with routine benefits and actual experience. For the selected situations $S_q$, the associated intent set is
\begin{equation}
\mathcal{I}_q=\bigcup_{s\in S_q}\mathcal{I}(s).
\label{eq:intent-selection}
\end{equation}
LiveAsset supplies all intents in $\mathcal{I}_q$ to the agent with their explanations, requirements, and facts, keeping the selected set fixed throughout the task while assessing the applicability of individual intents against the agent's actions and outcomes during execution.

\subsection{Runtime Intent Verification}
\label{sec:runtime}
An intent checker examines each agent output after its tool calls have taken effect and before the reply is delivered to the user. The checker uses the intent explanation to determine whether a learned intent applies to the current dialogue, actions, and outcomes. In Figure~\ref{fig:overview}, the draft states ``I save 10 min.'' The intent explanation connects personal claims in a product review to the user's actual experience, making this claim subject to verification. The stored fact says that the user has not timed this styler, so the claimed saving is unsupported. The learned requirement specifies using \texttt{[My time]} when timing is unknown. The checker therefore asks the agent to replace ``I save 10 min.'' with this placeholder while keeping the description of how combined drying and styling simplifies the morning routine.

The checker uses the current dialogue, reply, visible tool results, and files to judge each requirement and fact of every intent in $\mathcal{I}_q$ as \emph{met}, \emph{unmet}, or \emph{not applicable}. These judgments determine the intent status shown in Figure~\ref{fig:overview}: an intent is unmet if any requirement or applicable fact is unmet, not applicable if none of its requirements or facts apply, and met otherwise. Not-applicable judgments do not trigger revision. An unmet judgment explains how the current result fails to meet a requirement or conflicts with a recorded fact. Because later revisions can invalidate earlier fulfillment, LiveAsset rechecks every selected intent after each output, including those previously met.

If any selected intent receives a unmet judgment, LiveAsset asks the same agent to revise the result, providing the unmet requirements and facts with the checker's reasons, alongside requirements already judged met that should be preserved. The agent returns a complete revised result that replaces the previous reply before delivery to the user. Further revision stops when no unmet findings remain or a revision does not increase the number of requirements and facts judged met.

\section{Experiments}
\label{sec:evaluation}
\label{sec:results}
We evaluate LiveAsset across five benchmarks and two actor models to assess how effectively it learns reusable user intent from corrective interactions. Same-task replay measures fulfillment of requirements revealed during learning, while transfer evaluates whether the learned experience remains useful as requests and task contexts change. Ablations and further analyses examine the roles of interaction context and runtime correction, how relevant experience is selected and acted upon, and whether requirement fulfillment remains consistent across repeated executions.

\begin{table*}[t]
\caption{Performance on same-task replay and transfer across five benchmarks and two actor models. Satisfaction measures fulfillment of exposed user requirements; PROC incorporates delivery evidence, while TC and Reward measure task completion and trajectory quality. Bold marks the best result among Skill, Mem0, TellOnce, and LiveAsset.}
\label{tab:reuse}
\label{tab:transfer}
\label{tab:skillevol}
\centering\small
\setlength{\tabcolsep}{2pt}
\begin{tabular*}{\textwidth}{@{\extracolsep{\fill}}l*{6}{c}@{\hspace{10pt}}*{7}{c}@{}}
\toprule
& \multicolumn{6}{c}{\textbf{Same-task replay}} & \multicolumn{7}{c}{\textbf{Transfer to changed tasks}} \\
\cmidrule(lr){2-7}\cmidrule(lr){8-14}
& \multicolumn{2}{c}{$\pi$-Bench} & \multicolumn{2}{c}{PERMA} & \multicolumn{2}{c}{Vita} & \multicolumn{3}{c}{PAST-Bench} & \multicolumn{4}{c}{SkillEvolBench} \\
\cmidrule(lr){2-3}\cmidrule(lr){4-5}\cmidrule(lr){6-7}\cmidrule(lr){8-10}\cmidrule(lr){11-14}
Method & Satisfaction & PROC & Satisfaction & TC & Satisfaction & Reward & All & Near & Far & All & T4 & T5 & T6 \\
\midrule
\multicolumn{14}{l}{\textit{Actor: \mbox{GPT-6 Luna}}} \\
\midrule
Base agent & 0.218 & 0.466 & 0.132 & 0.114 & 0.228 & 0.015 & 0.641 & 0.648 & 0.634 & 0.815 & 0.809 & 0.840 & 0.794 \\
Raw history & 0.801 & 0.794 & 0.881 & 0.929 & 0.789 & 0.424 & 0.933 & 0.950 & 0.910 & 0.857 & 0.892 & 0.858 & 0.821 \\
\midrule
Skill & 0.554 & 0.681 & 0.704 & 0.657 & 0.733 & 0.545 & 0.836 & 0.839 & 0.833 & 0.854 & 0.882 & 0.855 & \textbf{0.826} \\
Mem0 & 0.607 & 0.722 & 0.848 & 0.771 & 0.689 & 0.576 & 0.898 & 0.892 & 0.904 & 0.832 & 0.844 & 0.871 & 0.782 \\
TellOnce & 0.738 & 0.807 & 0.882 & 0.743 & 0.728 & 0.424 & 0.749 & 0.764 & 0.729 & 0.806 & 0.796 & 0.846 & 0.775 \\
LiveAsset & \textbf{0.925} & \textbf{0.969} & \textbf{0.954} & \textbf{0.857} & \textbf{0.898} & \textbf{0.591} & \textbf{0.956} & \textbf{0.966} & \textbf{0.945} & \textbf{0.861} & \textbf{0.892} & \textbf{0.887} & 0.803 \\
\midrule
\multicolumn{14}{l}{\textit{Actor: DeepSeek V4.1-Flash}} \\
\midrule
Base agent & 0.321 & 0.537 & 0.108 & 0.100 & 0.378 & 0.067 & 0.601 & 0.640 & 0.551 & 0.824 & 0.814 & 0.837 & 0.819 \\
Raw history & 0.827 & 0.903 & 0.885 & 0.986 & 0.761 & 0.517 & 0.957 & 0.962 & 0.951 & 0.870 & 0.925 & 0.862 & 0.823 \\
\midrule
Skill & 0.713 & 0.780 & 0.863 & 0.786 & 0.756 & 0.583 & 0.841 & \textbf{0.866} & 0.810 & 0.867 & \textbf{0.906} & 0.863 & 0.832 \\
Mem0 & 0.530 & 0.645 & 0.814 & 0.686 & 0.633 & 0.467 & 0.821 & 0.852 & 0.781 & 0.861 & 0.888 & 0.877 & 0.818 \\
TellOnce & 0.806 & 0.809 & 0.848 & 0.771 & 0.928 & 0.567 & 0.803 & 0.777 & 0.837 & 0.858 & 0.876 & 0.878 & 0.821 \\
LiveAsset & \textbf{0.921} & \textbf{0.964} & \textbf{0.943} & \textbf{0.886} & \textbf{0.972} & \textbf{0.633} & \textbf{0.866} & 0.857 & \textbf{0.876} & \textbf{0.884} & \textbf{0.906} & \textbf{0.890} & \textbf{0.858} \\
\bottomrule
\end{tabular*}
\end{table*}

\subsection{Experimental Setup}
\label{sec:data}
\paragraph{Tasks and protocols.}
For same-task replay, $\pi$-Bench~\cite{zhang2026pibench} evaluates web research and professional reporting, with native checklists guiding our simulated user's corrections. PERMA~\cite{liu2026perma} evaluates personalized shopping and travel recommendation dialogues, where task goals and user preferences guide feedback. VitaBench 2.0 (Vita)~\cite{chen2026vitabench2} evaluates personalized online food ordering, in-store services, and travel booking through tool use, with trajectory rubrics guiding corrections.

For transfer, PAST-Bench~\cite{xue2026pastbench} evaluates experience reuse in note sharing, helpdesk, and knowledge-base workflows, with corrections added to designated learning episodes before near- and far-transfer tests. SkillEvolBench~\cite{lei2026skillevolbench} evaluates procedural reuse in API orchestration, document processing, and scheduling, using failed native checks to generate T1--T3 corrections before changed-scenario (T4), adversarial (T5), and composition (T6) tests.

We generate corrections using the user simulation in Section~\ref{sec:simulation}, based on current replies, tool observations, and files. Test targets start from their initial environments without further learning or simulated corrections, and Vita tasks are completed through its native service dialogue.

\paragraph{Compared methods.}
\label{sec:comparisons}
We compare LiveAsset with Base agent, Raw history, Skill, Mem0, and TellOnce. All methods that reuse experience learn from the same multi-turn corrective interactions. Base agent receives no information from these earlier interactions. Raw history supplies the complete recorded interactions at task entry, including user messages, agent replies, recorded tool results, and changed text files. Skill uses Codex\textquotesingle s \texttt{skill-creator} to turn the interactions into a reusable skill directory containing \texttt{SKILL.md} and any supporting files, then supplies the skill instructions and supporting text or code to the agent at task entry. Mem0 incrementally extracts and updates memories from the interactions, then retrieves up to five memories for each test-time user message and adds them to the agent's context. TellOnce compiles user feedback into persistent rules, applies rule-based validation after tool calls, and reviews compliance before delivery to guide the agent in correcting violations. LiveAsset incrementally learns intent assets, matches their situations to the target request, and checks the selected intents against the agent's actions and outcomes to trigger revisions for unmet requirements (Section~\ref{sec:method}).

\paragraph{Outcome measures.}
\label{sec:metrics}
Requirement satisfaction measures whether the agent's actions and final outcomes $y$ fulfill requirements revealed in the learning interactions $H_u$, following the objective in Equation~\ref{eq:objective}. In same-task replay, let $U_q$ contain the benchmark criteria corresponding to these requirements for the original request $q$. All methods are evaluated on the same set $U_q$. For nonempty $U_q$, the fraction of satisfied criteria, weighted by their benchmark weights $w_{qr}$, is
\begin{equation}
 \operatorname{Satisfaction}(q,y)=\frac{\sum_{r\in U_q}w_{qr}\mathbb{1}[y\text{ satisfies }r]}{\sum_{r\in U_q}w_{qr}}.
 \label{eq:satisfaction}
\end{equation}
Satisfaction is judged from the delivered reply, tool actions, and files using the native scoring criteria. $\pi$-Bench uses the item weights specified by the benchmark; PERMA and Vita assign weight 1 to every criterion.

\begin{figure}[!t]
\centering
\includegraphics[width=\columnwidth]{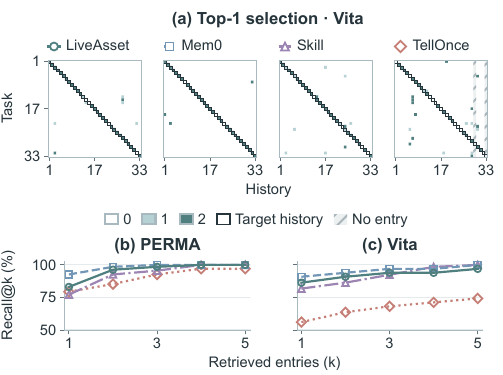}
\caption{Situation selection in same-task replay. (a) Top-1 selections on Vita, counted over two retrieval runs. (b, c) Recall@$k$ of each task's learning histories on PERMA and Vita.}
\label{fig:situation-selection}
\end{figure}

\paragraph{Models, harness, and budgets.}
We use two actor models, \mbox{GPT-6 Luna} and DeepSeek V4.1-Flash, both with maximum reasoning effort. Both models run through a common Codex harness with benchmark-specific tool adapters. Independent scoring uses \mbox{GPT-6.1 Sol} with high reasoning effort. In the current $\pi$-Bench comparison, asset construction and runtime checking use the corresponding actor with maximum reasoning effort. Methods share the actor, tools, initial environment, and actor budget within each setting.

\begin{figure*}[!t]
\centering
\includegraphics[width=0.87\textwidth]{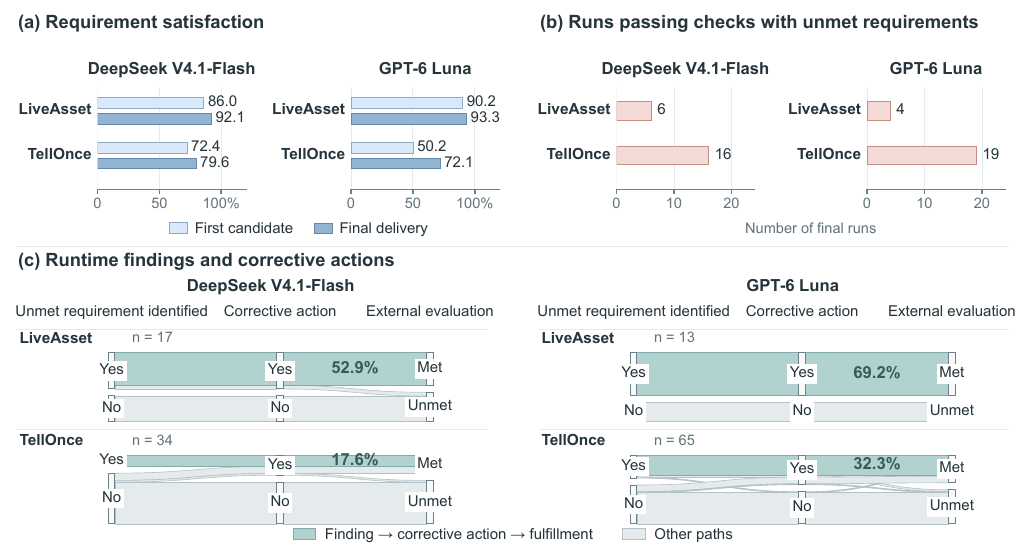}
\caption{Runtime verification and requirement fulfillment on $\pi$-Bench. (a) External satisfaction of the first candidate and final delivery. (b) Final runs that pass runtime checks but leave requirements unmet. (c) Paths of initially unmet requirements, shown as proportions of each method's $n$.}
\label{fig:behavior-alignment}
\end{figure*}

\subsection{Requirement Satisfaction in Replay}

Same-task replay revisits a request for which Raw history already contains the corrective feedback and its interaction context. LiveAsset achieves the highest requirement satisfaction on all three benchmarks with both actors (Table~\ref{tab:reuse}). On $\pi$-Bench's research and reporting tasks, LiveAsset also achieves higher PROC than Raw history.

PERMA evaluates shopping and travel recommendations in a single reply without tool use. LiveAsset's satisfaction exceeds Raw history by 7.3 percentage points with GPT-6 Luna and 5.8 with DeepSeek V4.1-Flash, while Raw history leads task completion. Vita extends the comparison to interactive ordering and booking with tool use, where LiveAsset also achieves higher native trajectory Reward than Raw history with both actors. The gains therefore extend from recommendation content to interactive service tasks.

\subsection{Transfer to Changed Tasks}

Transfer evaluates whether learned experience remains useful as task situations change. PAST-Bench tests near transfer in contexts similar to those seen during learning and far transfer under larger changes in topic, entities, or request wording. LiveAsset exceeds Skill, Mem0, and TellOnce in overall and far-transfer performance with both actors (Table~\ref{tab:transfer}). On far transfer, its advantage over the strongest of these baselines is 4.1 percentage points with GPT-6 Luna and 3.9 with DeepSeek V4.1-Flash.

SkillEvolBench tests this adaptation through changed scenarios (T4), adversarial follow-ups (T5), and composition (T6). LiveAsset outperforms Skill, Mem0, and TellOnce overall with both actors, and its T5 scores exceed the strongest of these baselines by 1.6 percentage points with GPT-6 Luna and 1.2 with DeepSeek V4.1-Flash. This shared advantage supports reusing learned requirements under adversarial follow-ups. On T6, Skill leads with GPT-6 Luna (0.826 versus LiveAsset's 0.803) and LiveAsset leads with DeepSeek V4.1-Flash.

\subsection{Analysis}
\label{sec:analysis}

\subsubsection{Ablation Study}
\label{sec:learning-ablation}
\label{sec:ablation}
Table~\ref{tab:ablation} examines intent explanations and corrective interaction during learning and execution on the three replay benchmarks. The variant without corrective interaction context keeps correction steps and earlier user messages but removes the surrounding agent replies, tool results, and changed files, testing their contribution to learning contextual intents. The variant without runtime correction reuses the full assets and task-entry matching but disables output checking and corrective continuation. Explanations are still provided, so this variant tests the runtime correction loop as a whole. The variant without intent explanations reuses the full assets and task-entry selections but removes explanations from the actor and checker inputs, keeping requirements, facts, and runtime correction unchanged, so it tests the role of explanations during execution and checking.

Full outperforms all three variants on satisfaction and task score across all three benchmarks and both actors. Removing intent explanations lowers satisfaction by 3.4--13.7 percentage points across these settings. On $\pi$-Bench, the decreases are 9.4 points with DeepSeek V4.1-Flash and 13.7 with GPT-6 Luna. With requirements, facts, and runtime correction held constant, these results support retaining explanations to guide execution and checking. The drop without corrective interaction context supports preserving the interaction context in which corrections occur during learning. With the same assets and task-entry matching, disabling runtime correction also lowers performance, supporting output checking and revision during execution. The context and runtime ablations reduce satisfaction most on Vita: removing context lowers it by 6.6 and 8.8 percentage points with DeepSeek V4.1-Flash and GPT-6 Luna, respectively; disabling runtime correction lowers it by 6.1 and 9.2 points.

\begin{table}[!t]
\caption{Ablation results for the full system and three variants on three replay benchmarks.}
\label{tab:learning-ablation}
\label{tab:ablation}
\centering\small
\setlength{\tabcolsep}{2pt}
\renewcommand{\arraystretch}{1.04}
\begin{tabularx}{\columnwidth}{@{}>{\raggedright\arraybackslash}X*{4}{>{\centering\arraybackslash}p{0.15\columnwidth}}@{}}
\toprule
Method / variant & \multicolumn{2}{c}{\textbf{DeepSeek V4.1-Flash}} & \multicolumn{2}{c}{\textbf{GPT-6 Luna}} \\
\cmidrule(lr){2-3}\cmidrule(lr){4-5}
 & Sat. & Task score & Sat. & Task score \\
\midrule
\multicolumn{5}{l}{\textit{$\pi$-Bench (PROC)}} \\
LiveAsset (Full) & \textbf{0.921} & \textbf{0.964} & \textbf{0.925} & \textbf{0.969} \\
\quad w/o explanation & 0.827 & 0.833 & 0.788 & 0.912 \\
\quad w/o context & 0.862 & 0.852 & 0.881 & 0.908 \\
\quad w/o runtime & 0.912 & 0.915 & 0.883 & 0.924 \\
\midrule
\multicolumn{5}{l}{\textit{PERMA (TC)}} \\
LiveAsset (Full) & \textbf{0.943} & \textbf{0.886} & \textbf{0.954} & \textbf{0.857} \\
\quad w/o explanation & 0.832 & 0.791 & 0.894 & 0.759 \\
\quad w/o context & 0.890 & 0.843 & 0.931 & 0.814 \\
\quad w/o runtime & 0.926 & 0.786 & 0.926 & 0.786 \\
\midrule
\multicolumn{5}{l}{\textit{Vita (Reward)}} \\
LiveAsset (Full) & \textbf{0.972} & \textbf{0.633} & \textbf{0.898} & \textbf{0.591} \\
\quad w/o explanation & 0.938 & 0.567 & 0.772 & 0.435 \\
\quad w/o context & 0.906 & 0.531 & 0.810 & 0.470 \\
\quad w/o runtime & 0.911 & 0.578 & 0.806 & 0.515 \\
\bottomrule
\end{tabularx}
\end{table}

\subsubsection{Situation Selection}
\label{sec:selection-analysis}

For task-entry retrieval in same-task replay, we pool artifacts learned with GPT-6 Luna by benchmark. We use only complete learning histories that include at least three agent deliveries, contain both corrective feedback and explicitly stated requirements, including user preferences, and end in user acceptance. We then select tasks meeting these conditions in descending order of the number of evaluation criteria. As a result, we evaluate 20 $\pi$-Bench, 34 PERMA, and 33 Vita tasks. The same model selects situations with associated intents for LiveAsset, complete skills for Skill, and active rules for TellOnce, while Mem0 uses native retrieval. Each task has two retrieval runs, each returning five entries without task execution. Recall@$k$ measures the fraction of a task's source histories represented among the first $k$ entries, averaging runs within each task and then tasks equally. TellOnce treats instructions from one PERMA and two Vita histories as task-specific and stores no rules for them.

Figure~\ref{fig:situation-selection}(a) shows first choices more concentrated on source histories for LiveAsset and Mem0. LiveAsset's advantage over whole-skill selection appears at Recall@1 (b,c), with 83.1\% versus 77.2\% on PERMA and 86.4\% versus 81.8\% on Vita, while Mem0 has the highest Recall@1 on both benchmarks. At Recall@5, Skill and Mem0 reach 100\% on both benchmarks, LiveAsset reaches 100\% on PERMA and 97.0\% on Vita, and TellOnce reaches 74.2\% on Vita. All $\pi$-Bench curves reach 100\% at Recall@1--5 and are omitted. This selection is thus competitive with skill and memory retrieval, and the transfer gains in Table~\ref{tab:transfer} suggest that it remains adequate for changed requests.

\begin{figure}[!t]
\centering
\includegraphics[width=\columnwidth]{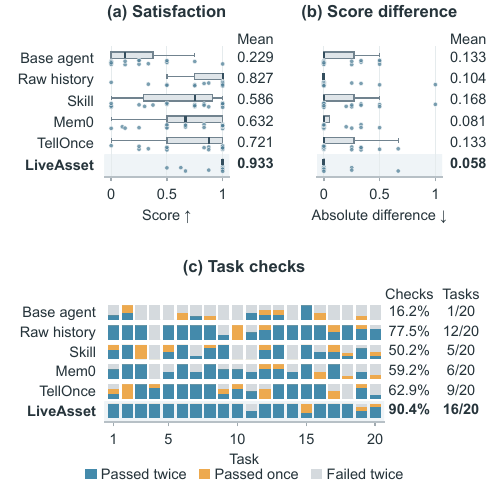}
\caption{Requirement fulfillment across repeated runs on $\pi$-Bench. (a) Per-task satisfaction. (b) Absolute score difference between two runs. (c) Per-task check outcomes, with the share of checks passed twice and the number of tasks passing all checks twice.}
\label{fig:repeat-stability}
\end{figure}

\subsubsection{Intent Fulfillment}
\label{sec:behavior-analysis}
\label{sec:application-repair}
\label{sec:unmet}

We use the 20 eligible $\pi$-Bench tasks with the most evaluation criteria (Section~\ref{sec:selection-analysis}) and compare LiveAsset with TellOnce under identical learning histories and exposed requirements (Figure~\ref{fig:behavior-alignment}). Each task is run twice per method, and the analysis covers 58 criteria corresponding to requirements exposed during learning.

LiveAsset's external satisfaction increases from 86.0\% to 92.1\% under DeepSeek V4.1-Flash and from 90.2\% to 93.3\% under GPT-6 Luna, exceeding TellOnce in both the first candidate and the final delivery (a). Fewer of its final runs pass runtime checks while leaving requirements unmet (b), with six versus 16 of 38 runs under DeepSeek V4.1-Flash and four versus 19 of 40 under GPT-6 Luna.

Panel (c) follows 129 initially unmet requirement--run records through runtime identification, corrective action, and external evaluation. The complete path covers 52.9\% of LiveAsset's records versus 17.6\% for TellOnce under DeepSeek V4.1-Flash, and 69.2\% versus 32.3\% under GPT-6 Luna. For LiveAsset, 18 of the 19 requirements identified by runtime checks are fulfilled after a corresponding corrective action, and most remaining unmet records are requirements the check does not identify. Together, these results show that LiveAsset's advantage arises both before and during runtime correction. It produces stronger first candidates (a), releases fewer deliveries with unmet requirements (b), and resolves most identified discrepancies through corresponding revisions that pass external evaluation (c).

\subsubsection{Requirement Fulfillment Across Repeated Runs}

On the same 20 $\pi$-Bench tasks, selected as in Section~\ref{sec:selection-analysis} and run twice with GPT-6 Luna, LiveAsset achieves higher requirement satisfaction than TellOnce, 0.933 versus 0.721, and the smallest mean absolute score difference between the two runs among all methods, 0.058 versus 0.133 for TellOnce (Figure~\ref{fig:repeat-stability}(a,b)).

Panel (c) further shows that more evaluated requirements are fulfilled in both runs. The mean share of checks passed twice is 90.4\% for LiveAsset and 62.9\% for TellOnce, and all evaluated checks pass in both runs on 16 of the 20 tasks, versus nine. LiveAsset's higher satisfaction is thus reproduced across executions, with most tasks fulfilling their full set of evaluated requirements in both runs.





\section{Conclusion}
We presented LiveAsset, an intent harness that learns user expectations from multi-turn corrective interactions to guide later agent execution. LiveAsset interprets corrections with original and revised results to incrementally build intent assets with explanations and checkable requirements, preserving earlier applicable requirements. The assets guide later tasks, while runtime verification uses explanations to assess applicability and requirements to check fulfillment and direct revisions before delivery. We formulated a corrective-interaction learning setting with simulated users and evaluated LiveAsset on five benchmarks covering same-task replay and cross-task transfer. Results show improved fulfillment of learned requirements and stronger performance on related tasks compared with memory, skill, and rule-based harness baselines. These findings support learning reusable user intent to guide execution and verify outcomes.

\clearpage
\bibliographystyle{ACM-Reference-Format}
\bibliography{references}
\clearpage
\appendix
\raggedbottom
\section{Interaction Evidence in Intent Assets}
\label{app:implementation}
\label{app:state}
Intent assets link each situation to excerpts from its source requests and each factual quotation to its source user message. We validate these excerpts and quotations against the user's original messages. When situations are merged, their source requests, associated intents, and source links are retained to preserve the context in which requirements and facts were learned.

\section{Learning Interactions and Feedback}
\label{app:learning-interactions}
\subsection{Analysis of User Feedback}
\label{app:feedback-analysis}
To examine how users express expectations for changes to the agent's work, we sampled 1,000 follow-up user-role messages without replacement from 41,211 records in SWE-chat~\cite{baumann2026swechat}, covering 720 coding sessions. After excluding progress, confirmation, new-task, derived or automatic records, and unresolved messages, we retained 331 corrective or clarifying messages from 285 sessions. All percentages in Figure~\ref{fig:correction-content} use these 331 messages as their denominator.

The Requirements group summarizes statements about expected outcomes and the context needed to interpret them. Preferences concern how work is performed or presented, task specifications state completion conditions, and context updates supply information needed to interpret these expectations. These disjoint subgroups contain 195 messages (58.9\%): 131 preferences (39.6\%), 55 task specifications (16.6\%), and nine context updates (2.7\%). Corrective feedback comprises 178 messages identifying a discrepancy in the work (53.8\%). Of these, 135 state no requirement (40.8\%), while 43 also express requirements (13.0\%). The two groups therefore overlap, and the dialogue excerpts illustrate how a message can identify a discrepancy while stating the desired outcome. One standalone execution-control message is omitted from the figure.

Annotations considered each message together with the preceding interaction, using additional context for ambiguous cases. Coding was assisted by a language model with the original dataset labels hidden. We reviewed ambiguous and long messages, residual categories, and random examples.

\subsection{Simulated Learning Interactions}
In $\pi$-Bench and Vita, native user turns handle preference information, clarification, and task interaction. When the native user would stop, the simulator reviews the work against the task criteria and adds corrections for unmet requirements; subsequent revisions return to the native user before another review. $\pi$-Bench uses checklists and tool checks, while Vita uses trajectory rubrics. PERMA reviews each delivered answer against task and preference goals, and SkillEvolBench uses failed native checks on T1--T3. PAST-Bench preserves the environment across baseline, learning, and reflection episodes, with task-outcome checks and corrections during the designated learning episodes. Interaction ends when the user accepts the result or the dialogue budget is exhausted.

Each review recomputes requirement fulfillment and passes met and unmet item identifiers to a stateful feedback selector. The selector records how often each requirement has been raised and which requirements were previously met. It favors unmet requirements raised least often, usually selects one concern or a few related concerns, and also includes previously met requirements that have become unmet. The simulator uses the selected requirements, interaction history, and current result to write the next user message in natural language. It points to a concrete discrepancy when available and otherwise states the desired change, expressing evaluation criteria as feedback on the work. Generated corrections are checked for grounding, coverage of the selected requirements, and added requirements before delivery.

User follow-ups in Table~\ref{tab:learning-interactions} include corrective feedback, answers to the agent's questions, and user replies needed to complete service tasks. Corrective messages count only feedback added after assessing the agent's work.

\begin{table}[t]
\caption{Mean user follow-ups ($\bar M$) and corrective messages ($\bar C$) per learning task or episode across five benchmarks.}
\label{tab:learning-interactions}
\centering\small
\setlength{\tabcolsep}{4pt}
\begin{tabular*}{\columnwidth}{@{\extracolsep{\fill}}lrr@{}}
\toprule
Benchmark & $\bar M$ & $\bar C$ \\
\midrule
$\pi$-Bench & 2.90 & 0.70 \\
PERMA & 2.11 & 2.09 \\
Vita & 4.27 & 0.85 \\
PAST-Bench & 1.83 & 1.83 \\
SkillEvolBench & 1.53 & 1.53 \\
\bottomrule
\end{tabular*}
\end{table}

\begin{table*}[t]
\caption{Benchmark coverage and evaluation measures for Table~\ref{tab:reuse}. Both actor models and all compared methods use the full evaluation scope listed below. Counts exclude repeated executions.}
\label{tab:benchmark-overview}
\centering\small
\setlength{\tabcolsep}{5pt}
\begin{tabularx}{\textwidth}{@{}p{0.13\textwidth}p{0.39\textwidth}X@{}}
\toprule
Benchmark & Learning and evaluation scope & Metric \\
\midrule
$\pi$-Bench & 100 tasks, 5 personas & Exposed checklist satisfaction; PROC with delivery evidence. \\
PERMA & 10 users, 580 user--query pairs; 1,094 checkpoints (705 single-domain, 389 multi-domain) & Exposed preference satisfaction; task completion (TC). \\
Vita & 56 users, 771 personalized subtasks & Exposed rubric satisfaction; native trajectory reward. \\
PAST-Bench & 26 families, 204 episodes; 34 near- and 26 far-transfer evaluation targets & Adapted task-outcome score, excluding prescribed memory/skill channels. \\
SkillEvolBench & 6 environments, 30 families, 180 tasks; T1--T3 learning and T4--T6 testing, 90 each & Native verifier reward and process/outcome checks. \\
\bottomrule
\end{tabularx}
\end{table*}

\section{Evaluation Details}
\label{app:protocol}
\subsection{Benchmarks and Measures}
\label{app:benchmarks}
Table~\ref{tab:benchmark-overview} reports the learning and evaluation scope used for the main results in Table~\ref{tab:reuse}. Same-task replay covers all listed tasks or user--query pairs; transfer covers the near- and far-transfer targets in PAST-Bench and T4--T6 in SkillEvolBench.

Requirement satisfaction is computed using Equation~\ref{eq:satisfaction}. PERMA task completion follows its official evaluation prompt, and Vita uses native trajectory reward. SkillEvolBench uses its native verifier reward. On PAST-Bench, task-outcome scoring removes credit for prescribed memory or skill channels and renormalizes the remaining task checks.

Delivery-evidence PROC evaluates $\pi$-Bench's hidden intents using the original targeted-follow-up classification, coverage formula, and eligibility rules. Its evidence includes the final reply, ordered AppWorld calls and results, final states, and delivered files. Work completed through tools receives credit unless an intent specifies an output channel.

\begin{table}[!t]
\caption{Mean token use for six methods across $\pi$-Bench, PERMA, and Vita, in thousands of tokens per execution (Test-time) or learning task (Build).}
\label{tab:efficiency}
\label{tab:luna-checking-cost}
\centering\small
\setlength{\tabcolsep}{2pt}
\begin{tabularx}{\columnwidth}{@{}Xrrrr@{}}
\toprule
& \multicolumn{2}{c}{DeepSeek V4.1-Flash} & \multicolumn{2}{c}{GPT-6 Luna} \\
\cmidrule(lr){2-3}\cmidrule(lr){4-5}
Method & \shortstack{Test-time\\(k/run)} & \shortstack{Build\\(k/task)} & \shortstack{Test-time\\(k/run)} & \shortstack{Build\\(k/task)} \\
\midrule
\textbf{LiveAsset} & 367.1 & 72.7 & 205.9 & 66.3 \\
Base agent & 243.6 & -- & 154.2 & -- \\
Raw history & 276.9 & -- & 293.0 & -- \\
Skill & 410.9 & 380.8 & 170.6 & 259.7 \\
Mem0 & 308.0 & 99.4 & 155.6 & 107.0 \\
TellOnce & 511.4 & 57.5 & 371.9 & 57.7 \\
\bottomrule
\end{tabularx}
\end{table}

\subsection{Execution Conditions and Aggregation}
We reset the target environment before each evaluation and keep learned assets fixed. $\pi$-Bench and PERMA produce one delivered reply. The former allows 240--540 tool steps and provides PDF and image readers; the latter uses no tools. Vita uses its service dialogue, with up to eight replies and 240 tool calls. SkillEvolBench permits 300 tool calls and a task-dependent execution time limit of 1,200 or 1,800 seconds. PAST-Bench permits 40 or 50 agent turns, depending on the task, with a 600-second time limit.

For each actor model, every method evaluates the full target set in Table~\ref{tab:benchmark-overview} twice. We first average the two executions within a target, then weight all evaluation targets equally.

For the ablation without intent explanations, the variant reuses the full system's frozen assets learned from the same histories. Requirements, facts, and their recorded scopes remain unchanged, and situation descriptions are used only for task-entry selection, which is identical to the full system. Both conditions share the actor, tools, initial environment, checker model and prompt template, checking schedule, revision and stopping rules, and execution and checking budget limits.

\begin{figure*}[t]
\centering
\includegraphics[width=\textwidth]{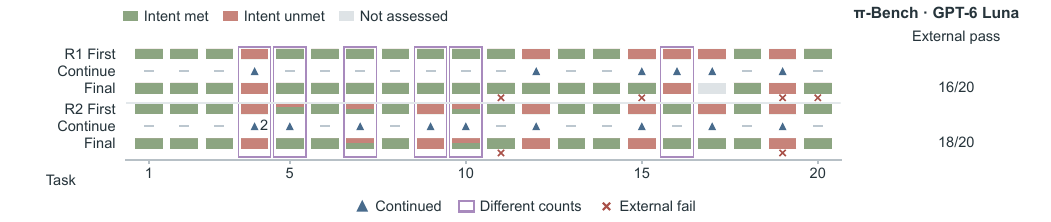}
\caption{Intent checks and revisions across two runs per task. Triangles mark continuation, with numerals for multiple revisions; purple outlines mark different revision counts. Crosses below final checks mark external failure; unmarked runs pass. External criteria follow Appendix~\ref{app:application-repair}.}
\label{fig:repeat-intent-paths}
\end{figure*}

\subsection{Computational Cost}
\label{app:ablation-cost}

Table~\ref{tab:efficiency} pools token use across the three same-task replay benchmarks. Construction cost is averaged over learning tasks, counting each asset's construction once. Test-time cost is averaged over executions and includes the agent and auxiliary model calls, including corrective continuation. It excludes the shared collection of learning interactions and independent evaluation. Token counts include input and output, with cached input and reasoning counted once. Dashes denote methods requiring no asset construction.

LiveAsset requires fewer construction tokens than Skill under both models, while its test-time cost is lower than Raw history under GPT-6 Luna and higher under DeepSeek V4.1-Flash.

\section{Additional Analyses}
\label{app:analyses}

\subsection{Experience Reuse}
\label{app:reuse-diagnostics}
The analyses in this section follow the task selection criteria described in Section~\ref{sec:selection-analysis}.

Skill satisfaction in Figure~\ref{fig:motivation}(a) averages two executions within a task and then weights tasks equally. The half violins use Gaussian density estimates with reflection at zero and one and a bandwidth of 0.10.

For memory retrieval, record coverage is the fraction of target records appearing among the top $k$ results. Targets are user memories present in the final store that were added or updated in subsequent learning interactions after the initial memories were formed. When a task has multiple source histories, we pool their target records. Coverage is computed per task and then averaged equally across tasks. Bands show pointwise 95\% percentile intervals from 2,000 task bootstrap resamples within each benchmark. Section~\ref{sec:selection-analysis} reports source-history recall for the same retrieval results.

The TellOnce analysis examines whether delivered results fulfill exposed requirements after the rule checker permits release. It includes 146 executions that pass rule-based checks and have complete external evaluations. Flow widths count executions; fulfillment is evaluated on the delivered results.

\subsection{Selective Reuse in Transfer}
\label{app:selective-reuse}
PAST-Bench traces illustrate selective reuse more directly. In a rule-migration task with GPT-6 Luna, the asset records the former review label as deprecated and explains that restricting note access must preserve a complete action summary. The agent uses the replacement label, retains every action item, and shares the note only with high-priority owners; the checker explicitly confirms that the deprecated label is absent and the sharing restriction is met. In a scoped-migration transfer task with the same actor, recipients are selected from the current note by the learned scope and priority rules, without carrying over historical recipient names. Both deliveries pass the external task-outcome checks, connecting contextual intent, runtime judgment, and observed actions in these cases.

\subsection{Intent Verification and Corrective Actions}
\label{app:application-repair}
Evaluation criteria include checklist items, tool-script checks, hidden-intent criteria, and a completion criterion. The analysis includes 58 criteria corresponding to requirements exposed during learning, with external judgments for both the first candidate and final delivery. Each requirement--run record pairs the first and final judgments of one criterion in one execution, giving 456 records.

First and final deliveries are assessed from their replies, tool traces, and files using the same native criteria. If all three remain unchanged, the initial and final judgments are shared. Internal verification passes when LiveAsset has no unmet applicable intent or TellOnce has no active violation. External fulfillment requires all criteria corresponding to exposed requirements to pass.

To examine how initially unmet requirements are subsequently handled, Figure~\ref{fig:behavior-alignment}(c) follows all 129 records whose first candidate fails the criterion, connecting checker feedback, corresponding corrective actions, and final judgments. A corresponding action must substantively change relevant reply content, files, or service state. The independent evaluation model annotates feedback and actions with native judgments hidden, and quotations are checked against the source interaction. Counts are normalized within each model--method group.

\raggedbottom
\noindent\begin{minipage}[t]{\columnwidth}
\subsection{Checks and Revisions Across Repeated Runs}
\label{app:repeat-intent}
Figure~\ref{fig:repeat-intent-paths} uses all 20 $\pi$-Bench tasks selected by the number of evaluation criteria for the main-text analyses, showing intent checks, revisions, and final outcomes for both runs of every task. Each column corresponds to one task. Bars show the proportions of met, unmet, and unassessed intents. Gray marks intents without a usable checker judgment. Intents judged not applicable are excluded when computing these proportions. A revision is a completed, nonempty agent output after a check, including text-only output.

The six outlined tasks have different revision counts across their two executions, yet pass externally in both. Repeated fulfillment can therefore involve different numbers of revisions.
\end{minipage}\par

\end{document}